\documentclass[lettersize,journal]{IEEEtran}
\usepackage{amsmath,amsfonts}
\usepackage{algorithmic}
\usepackage{algorithm}
\usepackage{array}
\usepackage[caption=false,font=normalsize,labelfont=sf,textfont=sf]{subfig}
\usepackage{textcomp}
\usepackage{stfloats}
\usepackage{url}
\usepackage{verbatim}
\usepackage{graphicx}

\usepackage{xcolor}
\usepackage{soul}
\usepackage{cite}
\begin{document}

\title{Distributed Learning and Inference Systems: A Networking Perspective}

\author{Hesham G. Moussa, Arashmid Akhavain, S. Maryam Hosseini, Bill McCormick}



\maketitle

\begin{abstract}
Machine learning models have achieved, and in some cases surpassed, human-level performance in various tasks, mainly through centralized training of static models and the use of large models stored in centralized clouds for inference. However, this centralized approach has several drawbacks, including privacy concerns, high storage demands, a single point of failure, and significant computing requirements. These challenges have driven interest in developing alternative decentralized and distributed methods for AI training and inference. Distribution introduces additional complexity, as it requires managing multiple moving parts. To address these complexities and fill a gap in the development of distributed AI systems, this work proposes a novel framework, Data and Dynamics-Aware Inference and Training Networks (DA-ITN). The different components of DA-ITN and their functions are explored, and the associated challenges and research areas are highlighted. 
\end{abstract}


\section{Introduction}
\IEEEPARstart{I}{n} recent years, human intelligence has been the reference for many researchers working in machine learning. Today we have models that exhibit similar or even surpass human capabilities in many fields including text (e.g., chatGPT), games (e.g., AlphaZero), and object recognition (e.g., YOLOv7) \cite{state_of_art}. While these results are impressive, the underlying model training paradigm relied mostly on centralization, where large volumes of data are collected and processed in a central cloud.

However,  centralization has become increasingly expensive, especially with the exponential growth in data volumes and model sizes \cite{centralized}. This cost is further aggravated in centralized life-long learning, which requires frequent initiation of costly re-training episodes to integrate new data without forgetting past learned tasks \cite{life_long_learning}. Additionally, transferring data from distributed nodes poses significant privacy and security risks.  Similar limitations also exist in centralized inference setups, especially when hosting large models on central servers which creates single points of failure, server congestion, and elevated computational cost. Furthermore, query and response data may carry sensitive information that, if exposed, may raise security and privacy concerns \cite{cloud_inf}.

To address these limitations, decentralization has gained considerable attention from researchers in recent years \cite{why_distributed}. Decentralized AI finds its way into many applications such as healthcare, robotics, and mobile networks \cite{wireless}. As such, finding ways to enable this paradigm has driven many innovations. Many decentralized training methods like distributed learning, federated learning, gossip learning, split learning, and continual learning \cite{gossib, distributed2}, along with decentralized inference approaches such as split inference, switched inference, collaborative inference, and multi-modal inference, have been proposed \cite{collaborative, inference_dist}.  These methods share the common objective of improving scalability, enhancing privacy, and reducing computational and storage demands \cite{distributed2}. This is achieved by distributing data and models across several networked nodes and enabling extensive mobility of the various components to enable AI functionality \cite{5g} — a concept known as the "model-follow-data" paradigm.

The model-follow-data paradigm views decentralized AI as a network of distributed, yet connected nodes. On the links between these nodes, models, data, and queries are optimally routed to facilitate AI training and inference. For instance,  in the case of model-follow-data-based training, the objective is to enable dynamic interactions between data, compute, and models for optimal model training by creating adequate rendezvous points in the network. For example, in vanilla federated learning (FL), distributed data nodes collaborate to train a global model. Each data node receives a copy of the model from a central server, trains it on its local data, and returns the updated weights to the server. The server aggregates these updated weights and creates an updated global model that is ready for the next round of training \cite{gossib}. Under vanilla FL, data nodes act as the rendezvous points for model-data interaction.

Similarly, in inference, the goal is to provide cheaper and faster query responses by optimizing the placement of models across the network to which queries can be routed efficiently. For instance, in 6G-enabled edge computing, servers near base stations bring computing close to the users; hence, by deploying AI models at these edge servers,  inference costs and response times can be significantly reduced \cite{edge}. In this setup, model-hosting facilities are considered the rendezvous points to which queries are routed which necessitates careful deployment designs.

Accordingly, under the model-follow-data paradigm, the success of training and inference relies on the careful design and optimization of several components. To address this requirement, we propose a novel network-inspired intelligent decentralized system that we refer to as Data and Dynamics aware Inference and Training Network (DA-ITN) which is the main contribution of this work.  DA-ITN aims to fill a gap in the next generation of distributed AI systems. In this paper, we dive into the general framework of DA-ITN and describe its various components. To the best of the authors' knowledge, this is the first work that introduces such a framework and vision.

The article is organized as follows. Section 2 provides an overview of DA-ITN and its various components. Section 3 dives into a high-level example to show  DA-ITN in action. Section 4 presents an envisioned forward-looking implementation of DA-ITN  as an autonomous system. The paper is wrapped up in section 5 with a list of challenges and potential research areas.

\section{DA-ITN overview}

DA-ITN aims to address decentralization complexities by: (i) understanding knowledge network topology, resource availability, and authorization; (ii) analyzing data characteristics like type, quality, and variability; (iii) assessing node resources (e.g., computing power, storage, energy) and heterogeneity; (iv) evaluating nodes for reachability, visibility, and trustworthiness; (v) selecting suitable model architectures (e.g., RNNs, CNNs, transformers) for inference or recommending model modification for better training performance; (vi) optimizing training hyperparameters at rendezvous points; (vii) deploying AI models for efficient inference; (viii) managing query-response mobility, and (ix) adapting to network connectivity conditions.

\begin{figure*}[!t]
\centering
\subfloat[]{\includegraphics[width=2.8in]{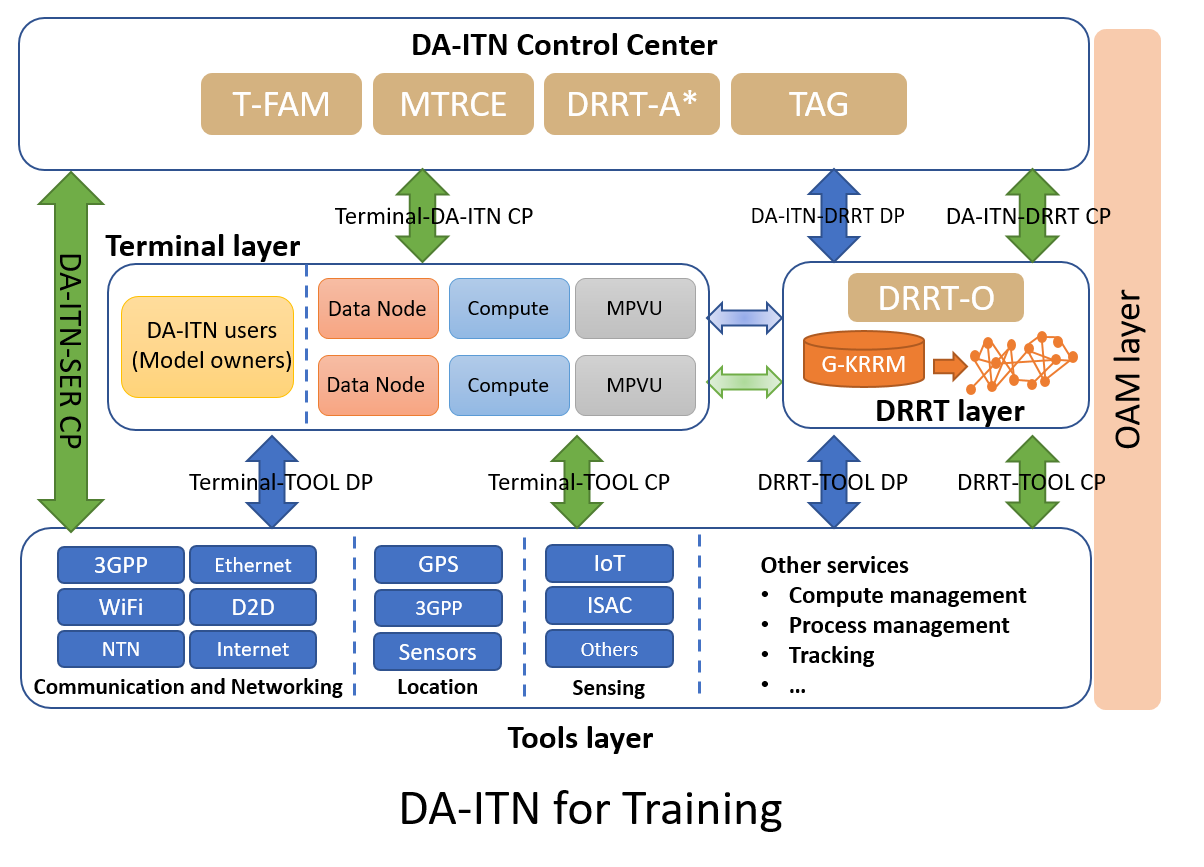}%
\label{fig_first_case}}
\hfil
\subfloat[]{\includegraphics[width=2.8in]{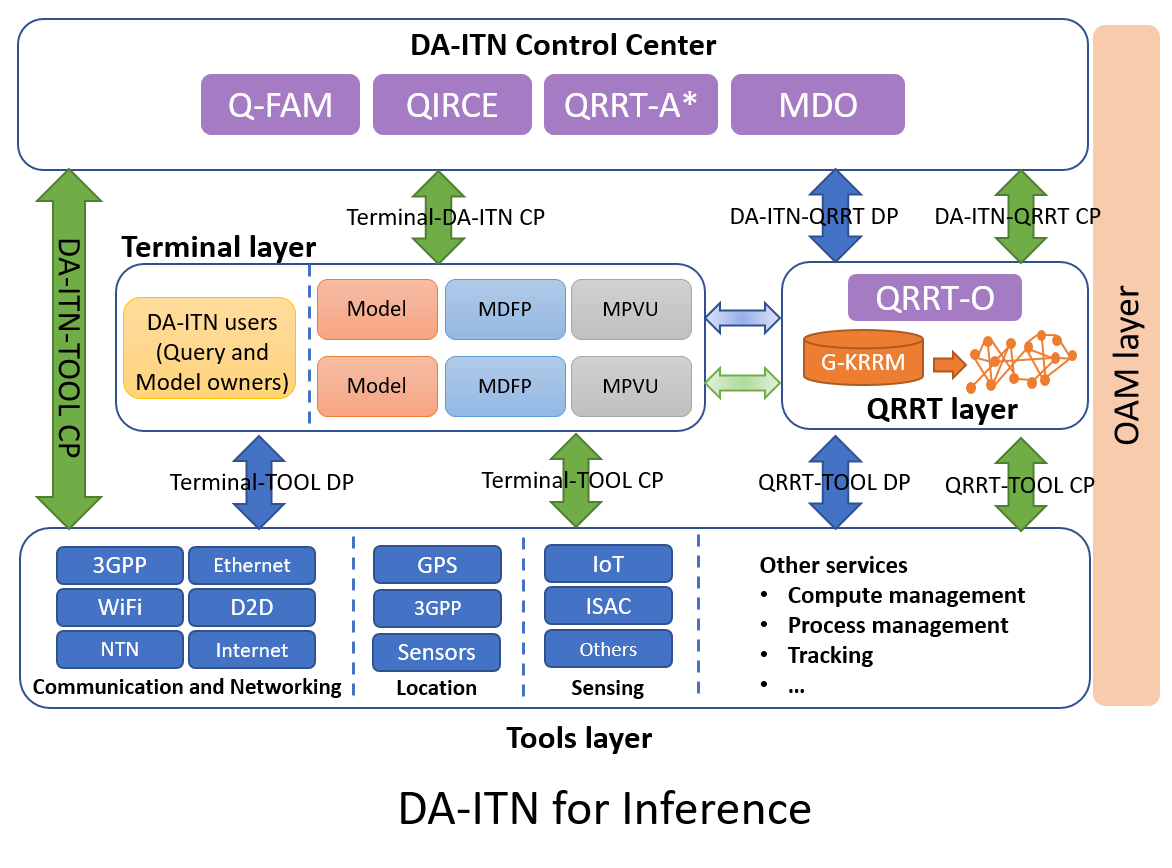}%
\label{fig_second_case}}
\caption{ (a) DA-ITN for training (DA-ITN-T). (b) DA-ITN for inference (DA-ITN-I).}
\label{fig_sim}
\end{figure*}

To handle these complexities, we envision DA-ITN as a network that contains a fully enabled control plane (CP), a data plane (DP), and an operations and management (OAM) plane as shown in Figure \ref{fig_sim}. The network uses its infrastructure to collect information about the participating parties' underlying data, resources, and reachability statuses, and creates knowledge topologies that aid in making intelligent AI training (Figure \ref{fig_first_case}) and inference (Figure \ref{fig_second_case}) decisions. In what follows, we discuss each of these DA-ITN setups. For the rest of the document, knowledge and data are used interchangeably unless otherwise stated.

\subsection{DA-ITN for Training}

DA-ITN for Training (DA-ITN-T) is a system designed to offer automated AI training services. It processes multiple users' training requests, models, and specific training requirements. Leveraging its underlying infrastructure, DA-ITN-T trains the submitted models accordingly. As illustrated in Figure \ref{fig_first_case}, the training architecture is structured into five layers: (i) the terminal layer, (ii) the tools layer, (iii) the Data, Resource, and Reachability Topology (DRRT) layer, (iv) the DA-ITN control center (DCC) layer, and (v) the OAM layer. These layers interact through control and data planes (CP and DP, respectively), as described below.

\textbf{The tools layer:} located at the core of the DA-ITN system, provides all the essential services that enable the functionality of DA-ITN-T. These include communication and networking, location services, sensing services, and compute and process management. All other layers rely on the tools layer, utilizing one or more of its services to fulfill their objectives. Notably, the communication and networking service tool from the tools layer establishes all the necessary CP and DP links, as depicted in the figure. This tool allows for the dynamic creation of adaptive CP and DP links to meet the needs of the terminal layer, which can grow or shrink as terminal components join or leave the network.  Each service in the tools layer may have a dedicated service manager, which other DA-ITN-T layers can connect to via CP links tailored to specific requirements. Alternatively, these service managers could be integrated into the other layers of DA-ITN-T, which help maintain the abstraction of the tools layer's services.

\textbf{The terminal layer:} consists of the system's terminal components. These include nodes that store the training data, facilities that provide the computing resources necessary for model training, and a newly introduced component known as the Model Performance Verification Units (MPVUs), where the model testing phase occurs. It is important to note that the facilities providing computing resources come in various forms: private property such as personal devices, distributed systems like mobile edge computing in 6G networks, cloud environments like AWS, or any other accessible location that can provide sufficient computational power for training, as long as both the data and the model can reach it. The MPVU plays a crucial role in distributed training by acting as a trusted proxy node. It holds a form of a constructed test dataset, which is built by collecting sample datasets from each participating node, and ensures secure and controlled access to it. Finally, the terminal layer also hosts the DA-ITN-T users, who are essentially the model owners seeking to utilize the training service. 

To interact with the tools layer, the terminal layer connects via the terminal-tool CP and DP planes, enabling the use of various services provided by the tools layer. For instance, the communication and networking service is used to build a tailored overarching knowledge-sharing communication network that helps: i) move models and data between compute points where training occurs, ii) transfer models to the MPVUs for performance evaluation to track training progress, and iii) allow DA-ITN-T users to submit models, monitor progress, modify training parameters, and retrieve trained models. This overarching network may utilize any access network technology including 3GPP cellular networks, WiFi, wireline, peer-to-peer, satellites, and non-terrestrial networks (NTN), or a combination of the above, to allow terminal components to access and exit the network and establish their trustworthiness. The network may also use transport layer protocols such as IP to establish global connectivity.  Beyond communication, the tools layer provides compute and process management services that can be leveraged by terminal compute nodes to support training. Additionally, sensing services can be used by data nodes to collect and store new data, potentially enhancing model training.

\textbf{The DRRT layer:} is a key element of the DA-ITN-T system. It serves as the bridge between the DCC and terminal layers, containing all the necessary information to support informed decision-making. It features a DRRT-orchestrator (DRRT-O) unit,  which connects to the tools and DCC layers via the DRRT-Tool CP and DA-ITN-DRRT CP links, respectively. The DRRT layer depends on various services from the tools layer to gather the data required for creating a Global Knowledge, Resource, and Reachability Map (G-KRRM), which is a large canvas that holds a high-level view of data, resources, and reachability information that is assumed to be in sync with the underlying knowledge network. Control commands for gathering this information are transmitted through the DRRT-tool CP link. Control commands for this data collection are sent via the DRRT-tool CP link and may either be handled by service managers in the tools layer or passed as configuration commands, enabling the DRRT-O to autonomously manage the tools layer services and hence help maintain abstraction.

\begin{figure}[!t]
\centering
\includegraphics[width=2.3in]{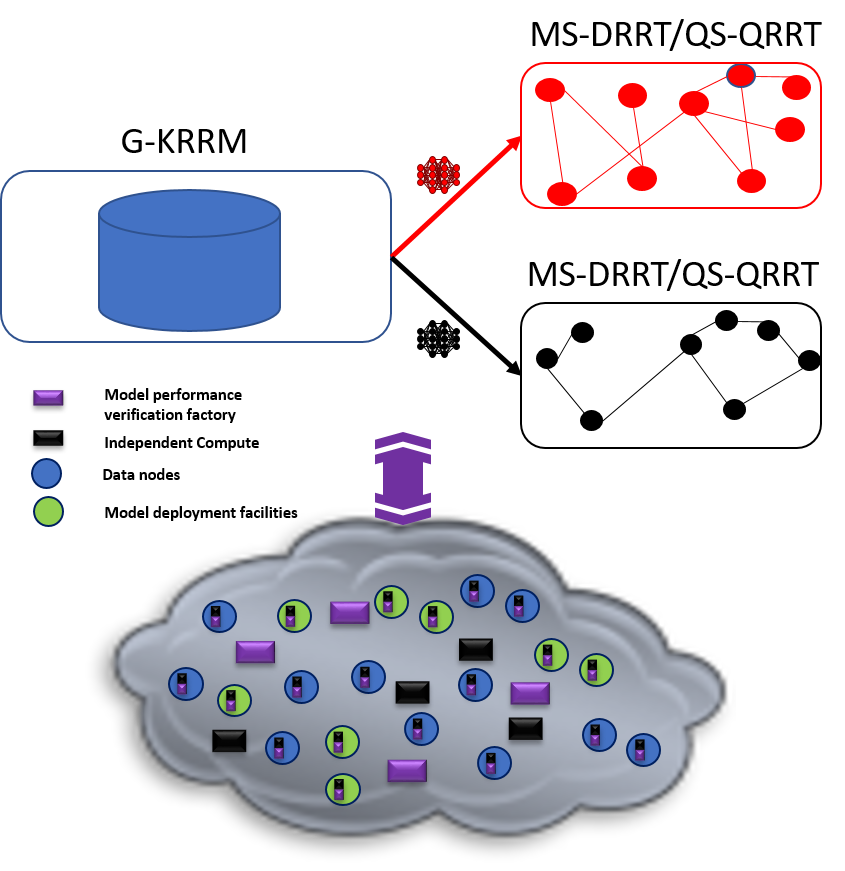}
\caption{Concept of the DRRT/QRRT topologies}
\label{DRRT}
\end{figure}

Additionally, the DRRT layer incorporates intelligence that enables it to transform the unstructured G-KRRM into Model-Specific structured DRRT topologies (MS-DRRT), which is a crucial step for decision-making within DA-ITN-T. These smaller, customized topologies help minimize computational costs and expedite decision-making. Figure \ref{DRRT} shows an example of this process.

The model-specific DRRT topologies hold information regarding the type, quality, volume, age, dynamics, and any other essential information about the data available for training a specific model. It also provides information on the reachability of the participating nodes to avoid unnecessary communication overhead or loss. They also include data about available computing resources and MPVUs, such as resource availability, location, trustworthiness, and details about testing datasets hosted at various MPVUs. The primary clients of these topologies are components in the DCC layer, as outlined next.

\textbf{The DA-ITN control center (DCC) layer:} is the topmost layer shown in the figure, and it houses the intelligence needed to make critical decisions based on DA-ITN user requirements. It includes components such as the Model Training Route Compute Engine (MTRCE), which determines where model-data rendezvous should occur, and the Training Feasibility Assessment Module (T-FAM), which evaluates whether training is feasible based on the submitted model, training requirements, and the state of the underlying knowledge sharing network. Additionally, the DCC layer contains intelligent modules like the Training Algorithm Generator (TAG) and the Hyper-Parameter Optimizer (HPO), responsible for decisions related to training methods (e.g., RL, FL, SL), number of epochs, batch size, and other training parameters.

The DCC layer connects to the tools layer through the DA-ITN-tool CP link, which transmits control data from terminal users, including training instructions, model structure, training requirements (e.g., accuracy, convergence time), progress monitoring requests, and configuration modifications. It also carries feedback to support components like TAG and HPO in making informed training decisions.

The DCC layer also links to the DRRT layer via the DA-ITN-DRRT CP and DP links, which provide essential information about the terminal layer components, enabling accurate decision-making. This information is delivered by the DRRT-O unit as model-specific DRRT topologies, which assist the MTRCE in making routing decisions and the T-FAM in model admission. An optional component within the DCC (or potentially part of the DRRT layer) is the DRRT-Adaptability Unit (DRRT-A), responsible for monitoring and updating model-specific topologies as training progresses and the terminal layer evolves. The DRRT-A is optional as the DRRT-O could take on this role by periodically updating the topologies based on specific triggers.

\textbf{The OAM layer:} spans across all layers, serving as a management layer to configure DA-ITN-T components, manage network connectivity, and enable feedback functions crucial for progress monitoring and model tracking. It also provides ongoing feedback to DA-ITN users about their models throughout the training process.

\subsection{DA-ITN for Inference}
Similar to DA-ITN-T, DA-ITN for Inference (DA-ITN-I) is a system that provides automated AI inference services using a similar infrastructure with a few differences as shown in Figure \ref{fig_second_case} and as discussed in the following.

First, unlike DA-ITN-T, where the moving components are models and training data and the rendezvous points are compute facilities, in DA-ITN-I, models, and queries are the moving components that require networking, and the rendezvous points are model hosting facilities. 

Second, in the  DA-ITN-I users are query and model owners. Query owners are the inference service users who send their queries into the system and collect the resulting inference. On the other hand, model owners are divided into two types. The first type consists of model hosts - the model used for inference does not have to be owned by them, but it is hosted on their computing facilities.  The second type consists of model providers - they develop models and deploy them either at their own facilities or at model hosts. Model owners are represented in the terminal layer as model deployment facility providers (MDFP) which are distributed across the global network.

Third,  \textbf{the tools layer} for inference provides the following services to the terminal layer using its control and data planes: i) model mobility from model generators to model hosts; ii) query routing towards models deployed on MDFPs; iii) model mobility from one location to the other in case of load balancing situations; iv) model mobility towards re-training and calibration facilities which may be hosted on MVPF units (common with  DA-ITN-T);  v) query response and inference result routing towards the query owners or any indicated destination around the globe; and vi) feedback and monitoring information to model and query owners.

Fourth, the  \textbf{Query, Resource, and Reachability Topology  (QRRT) layer} replaces the DRRT layer and provides similar services, but with a focus on models and queries. It offers information on both models and queries, including: i) for models: their locations, capabilities, current load conditions, inference speed, accuracy, reachability, and accessibility (specifically, the reachability and accessibility of the MDFP); and ii) for queries: query patterns, dynamics (potentially linked to geographic locations), query types, and the reachability status of query owners for response communication. This collected information is used to inform decisions about model deployment and distribution, query-to-model routing, and response routing. The QRRT layer also includes an orchestration function that collaborates with the tools layer to gather the necessary data from the terminal layer. This data is used to build the G-KRRM, which is then converted into Query-Specific topologies (QS-QRRT). Additionally, the QRRT layer may optionally interact with the QRRT-adaptation (QRRT-A) component to ensure these topologies are continuously updated and aligned with real-world changes.

Last, \textbf{the DCC layer}  contains several intelligent decision-making components, including the Query Feasibility Assessment Module (Q-FAM), the Query Inference Route Compute Engine (QIRCE), and the Model Deployment Optimizer (MDO). These components, similar to the training decision-making units, rely on the QRRT for their operations. The Q-FAM serves as an admission control unit, assessing whether a submitted query can be serviced based on the current network's inference capabilities. The QIRCE is responsible for routing queries to the appropriate models while considering load conditions. Meanwhile, the MDO functions as an admission controller for new models, evaluating their deployment feasibility based on architecture, compute, and storage requirements. It matches these needs with the available resources in the QRRT and makes a deployment decision, also optimizing the model's deployment location to reduce query response times and inference costs.

\subsection{DA-ITN as a Network}

\begin{figure*}[!t]
\centering
\includegraphics[width=5.4in]{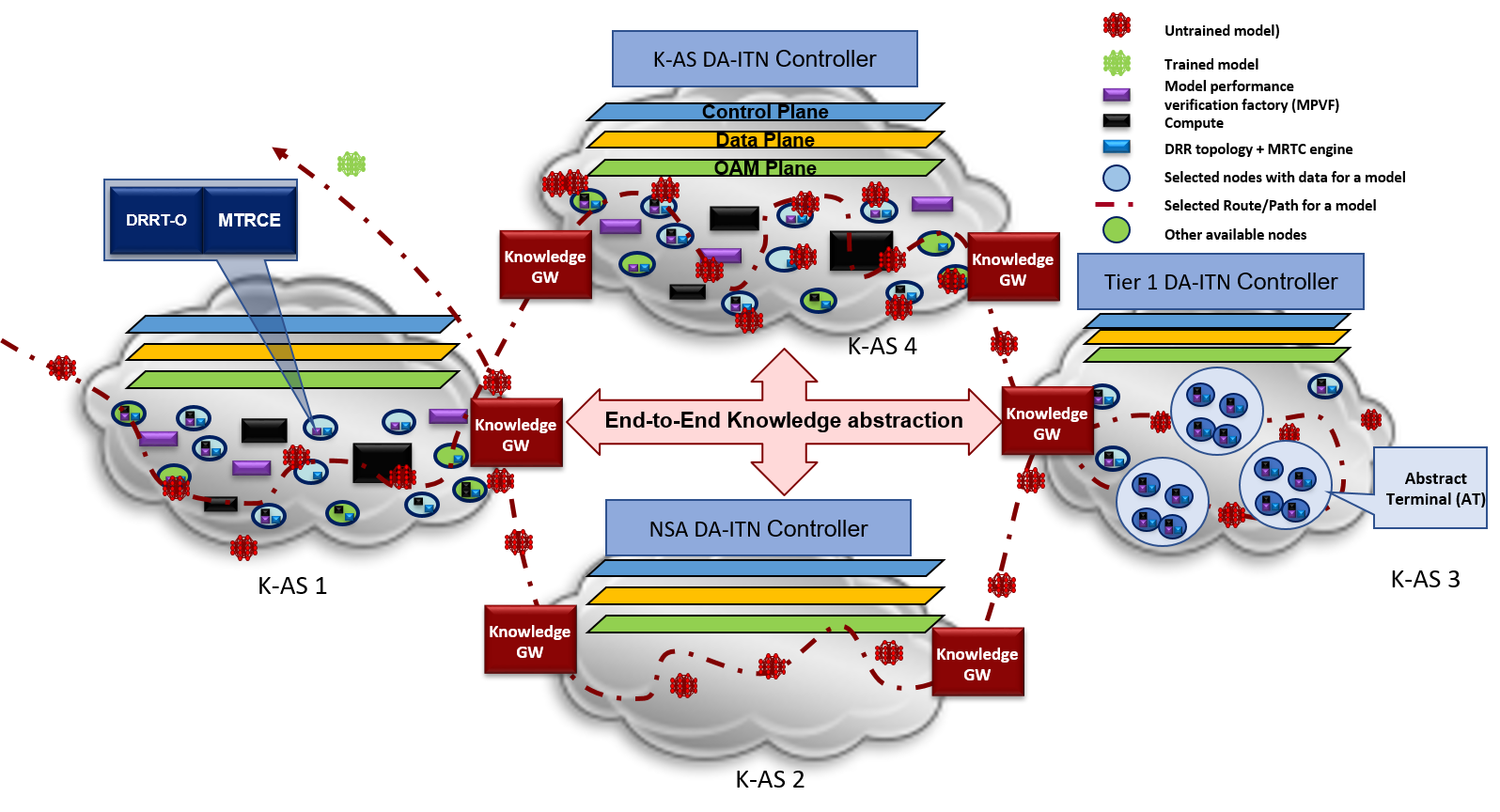}
\caption{Hirarichal setup for DA-ITN}
\label{fig_1}
\end{figure*}

At its core, DA-ITN, as a network, represents a novel type of system with components analogous to those of existing networking technologies, but specifically designed for DA-ITN. It introduces distinct elements within its control, data, and OAM plane (e.g, DRRT/QRRT, T-FAM/Q-FAM, and MTRCE/QIRCE). The functions of DA-ITN can be implemented either centrally or in a distributed manner, particularly when utilizing concepts like abstraction and hierarchy. Figure \ref{fig_1} illustrates an end-to-end DA-ITN network with various possible implementations.

As depicted in the figure, the global knowledge-sharing network, consisting of various terminal nodes (data nodes, compute nodes, MPVU nodes, and MDFP nodes), is divided into sub-networks known as Knowledge Autonomous Systems (K-AS). Each K-AS consists of a set of adjacent terminal components that form the local terminal layer. The combined local terminal layers from all K-AS regions together make up the overall end-to-end terminal layer of DA-ITN. Moreover, these K-AS regions are interconnected by knowledge border gateways that use a form of hierarchical communication protocols to support end-to-end DA-ITN services.

It is important to note that the local DA-ITN functions can be implemented in different ways. For example, the K-AS 1 region may implement a fully distributed DA-ITN system, where each terminal component may house some or all of the intelligence typically found in the DA-ITN control center. A data node, for instance, might run a DRRT-O that utilizes the local CP and DP links within its K-AS to gather data necessary for building a local DRRT topology. It could also host an MTRCE that uses this locally generated topology to make mobility decisions, such as determining the next hop for a model currently being trained on local data. On the other hand, a region like K-AS 4 could adopt a fully centralized DA-ITN system for making centralized mobility decisions within its intelligent, isolated local terminal layer.

Furthermore, DA-ITN functions may follow a hierarchical structure, as seen in K-AS 3. In this setup, a tier-1 DA-ITN control center oversees mobility decisions for the entire K-AS region. Meanwhile, abstract terminals (ATs)—which consist of one or more terminal components grouped but viewed as a single terminal component by the network—may have their own tier-2 DA-ITN control centers to handle local mobility decisions within the AT. These tiered DA-ITN control centers work together to enable K-AS-wide mobility decision-making. It is also important to note that the intelligence for mobility decision-making within a AT could differ from the DA-ITN system described here.

It is important to introduce the concept of a non-standalone DA-ITN control center, as shown in K-AS 2. This refers to a control center that does not possess all the necessary intelligence on its own and may rely on third-party assistance to make accurate decisions. For example, a non-standalone DA-ITN control center may host an MTRCE but lack a T-FAM, with the T-FAM functionalities provided by other DA-ITN control centers within different K-AS regions. Notably, this non-standalone control center can exist in centralized, decentralized, or hierarchical configurations using similar concepts as those discussed.

\section{DA-ITN in Action}
In this section, we look into sequential model training in healthcare \cite{sequential} as a use case for DA-ITN-T and describe a step-by-step process starting with model submission.

Imagine country A is planning to deploy department-specific LLM-based assistant AI models to be used by doctors to help speed up medical diagnoses. Each department-specific AI model is assumed to be an expert in a specific field with a broad enough knowledge of other medical fields. To train such an AI model, data from all hospitals and medical centers in the country is used to emphasize the area of intended expertise. However, due to the medical data's size and privacy restrictions, collecting it in a central location is not an option. Sequential learning, a distributed learning paradigm, is chosen to train the AI models instead, where data is assumed to be distributed across multiple nodes, and a single copy of the model is moved between the nodes for local training. The final performance of the model depends on the optimal choice of the training sequence, which depends on the model structure, the nature of the training data, and the training hyperparameters.

In this scenario, DA-ITN-T can be used to train these models. The first step is to collect all the necessary information which includes information about the medical data hosted at each healthcare facility in the country, information about the available computing resources, their accessibility status as well as their trustworthiness scores. To collect this information the DRRT layer coordinates with the service layer over the DRRT-SER CP link. The service layer then coordinates with the terminal layer to facilitate the collection of the necessary information required by the DRRT layer which would then build an accurate global data, resource, and reachability topology. 

The different AI models and their respective training requirements are then submitted to the DA-ITN Control center. For instance, an AI model intended to help the cardiology department is required to have broad medical knowledge with an accuracy of at least 80\% and expert cardiology knowledge with an accuracy of at least 95\%. The T-FAM within the DA-ITN control center then uses this information to assess the feasibility of the service. It may use the DRRT service provided by the DRRT layer to gather the data needed for service admission decisions.

For accepted models, the MTRCE collaborates with the DRRT-A unit to generate model-specific DRRTs. These tailored topologies are used to determine the optimal sequence of nodes the AI model should visit to achieve its objective. The MTRCE also sets the training hyperparameters and decides the number of times a model should visit an MPVF unit for performance assessment. The MTRCE's decisions are communicated to the communication services provided by the service layer to handle model mobility according to the specified sequence. Once training is complete, the models are returned to the owners, along with the training logs, and are ready for deployment.

\section{Envisioned Fully Autonomous DA-ITN}
\begin{figure}[!t]
\centering
\includegraphics[width=2.8in]{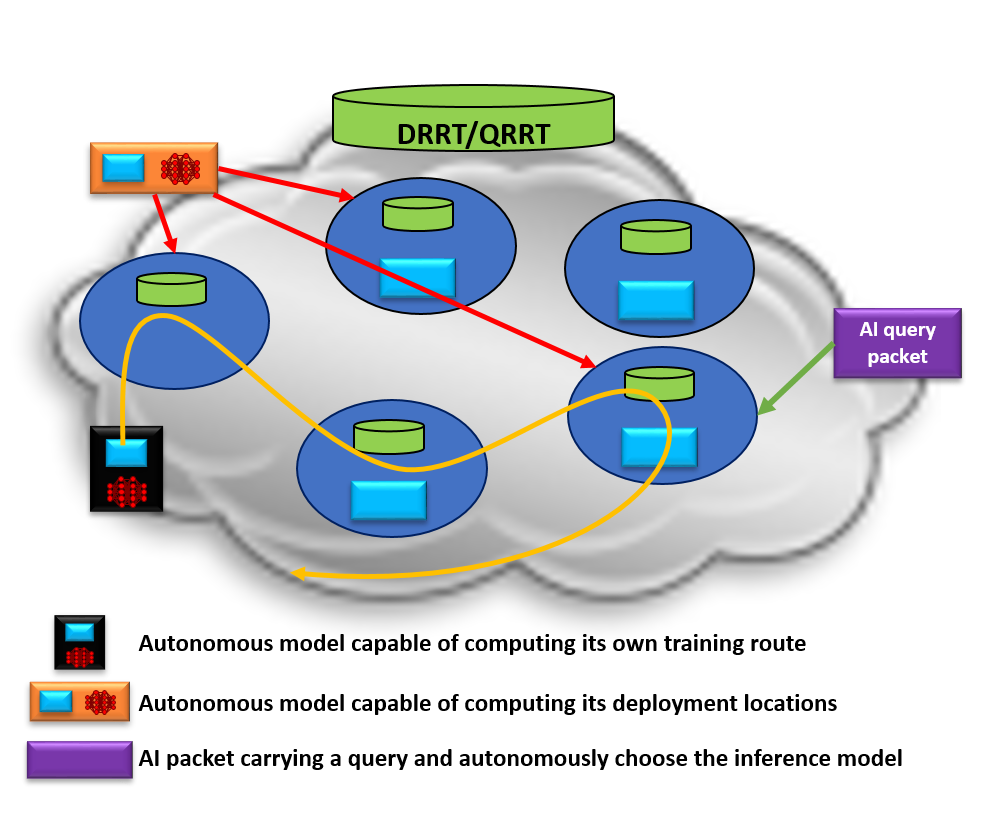}
\caption{Fully Autonomous DA-ITN with various AI objects}
\label{fig_3}
\end{figure}

Having outlined the general concept and components of the proposed DA-ITN, this section explores a forward-looking potential deployment setup. We imagine a fully autonomous system capable of learning the various functions of DA-ITN with a minimal architectural setup. Unlike the previously discussed network-like implementation, this envisioned system introduces intelligent autonomous entities (AI objects) that independently navigate the network. AI objects have algorithmic intelligence and attach themselves to nodes in the knowledge networks to consume network information, allowing them to autonomously steer themselves without relying on centralized control as illustrated in Figure \ref{fig_3}. The intelligence is embedded within these AI objects, enabling them to use self-contained information-gathering methods to make precise steering decisions and achieve specific objectives, such as training, inference, or model deployment. 

Similarly, we envision AI objects as unique self-operated network objects that can gather local and network-wide data, resource, and reachability information and independently make micro and macro AI-specific traffic steering decisions.  This setup is called the Autonomous AI Traffic Steering (AATS) framework. The main identifying unique feature of the AI objects of the AATS over typical communication network objects is the fact that they do not contain a destination address, they rather compute the destination based on the requirements of the payload as well as the data, resource, and reachability information gathered about the terminal nodes in the network.

For instance, under AATS, an AI packet that carries a model that was submitted for training by client A has a header that carries only the address of client A, but no destination address. The AI packet autonomously determines its destination based on available resources, training requirements, and network state. Initially, it gathers information from terminal nodes, and uses this information along with the payload information about the model to select destinations, and set destination addresses accordingly.  The AI packet then may use regular networking protocols to move toward the indicated destinations. The next address computation process could happen after each visit or could be computed end-to-end from the beginning.  For inference queries, the packet identifies the appropriate model and routes itself to it with the goal of meeting speed and accuracy requirements. For AI objects carrying generated models that are to be deployed, they autonomously determine the destined MDFPs and route themselves toward them.

\section{Challenges and Research Directions}
\subsection{DRRT and QRRT Generation}

The DRRT and QRRT topologies are vital for enabling intelligent decision-making in DA-ITN but remain challenging to develop due to their complexity and current nonexistence. Below, we list some of these challenges.

\textbf{Definition and Complexity:} In Figure \ref{DRRT}, we used cyclic graphs to visualize a DRRT, however, these topologies have much higher levels of complexity that go beyond a simple graph. DRRT and QRRT are envisioned as dynamic, map-like structures capable of addressing complex queries, such as fulfilling training requests based on model requirements, network resources, and node states. These topologies must capture multi-dimensional relationships beyond simple graph representations, requiring a deep understanding and novel construction methods.

\textbf{Data Overhead:} Constructing these topologies demands collecting extensive information about data from the terminal layer, potentially overburdening the network. Innovative techniques are needed to build sophisticated topologies while minimizing data requirements.

\textbf{Privacy Concerns:} While decentralization enhances privacy, gathering terminal-layer data introduces security risks. Developing methods to disguise sensitive data, such as using generative AI for secure representation, is crucial.

\textbf{Real-Time Synchronization:} The topologies must remain synchronized with the terminal layer's state, necessitating close to real-time communication and processing capabilities that are currently insufficient, as per digital twin research  \cite{DT_paper}.

Further research is essential to address these challenges and realize the potential of DRRT and QRRT topologies.

\subsection{ DA-ITN Control Center Intelligence}

Unlike the challenges of topology generation, building intelligence within the DA-ITN control center is a more attainable goal. Many proposed functions for the control center are already explored in existing literature. For example, node selection aligns with research in federated and split learning, while AI model design relates to neural architecture search (NAS) and hyper-parameter optimization\cite{NAS}. However, two key challenges remain: i) developing a framework for these methods to work synergistically toward DA-ITN's goals, and ii) addressing privacy concerns, as many approaches rely on access to physical data; hence, novel privacy-forward solutions are therefore essential.

On the other hand, intelligent components such as the T-FAM and Q-FAM are newly envisioned components that might benefit from some works available in the literature. However, dedicated solutions and methodologies are needed for these components to bear fruit and achieve their intended objectives.

Lastly, while we have mentioned that DA-ITN can be implemented in centralized, distributed, or hierarchical forms, we have not yet delved into the specific implementation strategies for each, which undoubtedly present a range of challenges. While a centralized implementation may be relatively straightforward, both distributed and hierarchical approaches introduce multiple layers of complexity that need to be addressed, particularly in terms of communication, abstraction, and decision-making. Achieving the behavior of a centralized network in a distributed system is still far from feasible with current technology.

\section{Conclusion}
In this work, a new framework to enable data, resource, and reachability-aware AI traffic steering system that provides optimized automated training and inference as a service is proposed. The various components of the proposed system are described and their functions are detailed. A use case was presented to show the system in action. The work was concluded with a list of potential challenges and areas for research and improvement. 

The introduced networks are specifically designed to enable efficient distributed AI training and inference. To become viable, such networks have a unqiue set of challenges that necessitate extensive fundamental research in data explainability, AI training and inference, and their associated algorithms. 

\bibliographystyle{IEEEtran}
\bibliography{References.bib}

\section*{Biography Section}

\vskip -2.5\baselineskip plus -1fil
\begin{IEEEbiographynophoto}{Hesham G. Moussa}
 received his B.S. and M.S. degrees from the American University of Sharjah, Sharjah, UAE, and his Ph.D. degree from the Department of Electrical and Computer Engineering, University of Waterloo, Waterloo, ON, Canada in 2020. He is currently working as a senior research engineer with the wireless department at Huawei Technologies Canada. His research interests include machine learning and its applications in wireless communications, and performance optimization for mobile networks.
\end{IEEEbiographynophoto}
\vskip -2.5\baselineskip plus -1fil

\begin{IEEEbiographynophoto}{Arashmid Akhavain}
is a network and system architect, a research engineer, and the leader of the Advanced Networking Research team under the Wireless Systems Division at Huawei Technologies Canada. He received his degree in Computer Engineering from Concordia University, Montreal. He has over 30 years of extensive design experience in a diverse set of technologies ranging from Telephony, Ethernet, and IP/MPLS/Segment-Routing, to VPN and Traffic Engineering, to SDN, NFV, and Mobile Networks. He worked for multiple major companies including Bell Northern Research, Nortel, and Ciena, where he held multiple leading roles in numerous projects.  He has contributed to several standards submissions. He is a prolific patent contributor and is currently focusing his efforts on advanced networking and AI/ML research and their inclusion and impact on 6G networks at Huawei.
\end{IEEEbiographynophoto}
\vskip -2.5\baselineskip plus -1fil
\begin{IEEEbiographynophoto}{S. Maryam Hosseini}
received a B.Sc. degree in electronics engineering from the University of Tehran, Tehran, Iran, in 2013, an M.S. degree in electrical engineering from Iran University of Science and Technology (IUST), Tehran, Iran, in 2015, and a Ph.D. degree in AI and machine learning from University of Waterloo, Waterloo, ON, Canada in 2023. She worked at Huawei Ottawa Research and Development Centre, Ottawa, ON, Canada as an intern from February 2023 to Dec 2023. She is currently a research engineer at Huawei Ottawa Research and Development Centre, Ottawa, ON, Canada. Her current research interests include machine learning and its applications in wireless communications. 

\end{IEEEbiographynophoto}
\vskip -2.5\baselineskip plus -1fil
\begin{IEEEbiographynophoto}{Bill McCormick}
is a principal engineer with Huawei Technologies Canada, where he is responsible for the development of practical applications of neural networks.   His research interests include distributed and self-organizing systems, as well as optimization and machine learning. McCormick received his master’s degree in electrical engineering from Carleton University.
\end{IEEEbiographynophoto}

\vfill

\end{document}